\documentclass[english]{cccconf}
\usepackage[comma,numbers,square,sort&compress]{natbib}
\usepackage{epstopdf}
\usepackage{amsmath}
\usepackage{amssymb} % 引入 amssymb 宏包
\usepackage{hyperref}
\usepackage[capitalise]{cleveref}
\usepackage{booktabs}
\usepackage{algorithm}
\usepackage{algpseudocode}
\usepackage{graphicx}  % 需要在导言区引入
\usepackage{tabularx}
\usepackage{booktabs}
\usepackage{multirow}
\begin{document}

\title{Hybrid A* Path Planning with Multi-Modal Motion Extension for Four-Wheel Steering Mobile Robots}

\author{Runjiao Bao\aref{bit},
        Lin Zhang\aref{bit},
        Tianwei Niu\aref{bit},
        Haoyu Yuan\aref{bit},
        Shoukun Wang\aref{bit}}

\affiliation[bit]{School of Automation, Beijing Institute of Technology, Beijing 100081, China; E-mail: 3120230765@bit.edu.cn}

% \affiliation[bit2]{Key Laboratory of Servo Motion System Drive and Control, Ministry of Industry and Information Technology, School of Automation, Beijing Institute of Technology, Beijing 100081, China}
\maketitle

\begin{abstract}
Four-wheel independent steering (4WIS) systems provide mobile robots with a rich set of motion modes, such as Ackermann steering, lateral steering, and parallel movement, offering superior maneuverability in constrained environments. However, existing path planning methods generally assume a single kinematic model and thus fail to fully exploit the multi-modal capabilities of 4WIS platforms. To address this limitation, we propose an extended Hybrid A* framework that operates in a four-dimensional state space incorporating both spatial states and motion modes. Within this framework, we design multi-modal Reeds-Shepp curves tailored to the distinct kinematic constraints of each motion mode, develop an enhanced heuristic function that accounts for mode-switching costs, and introduce a terminal connection strategy with intelligent mode selection to ensure smooth transitions between different steering patterns. The proposed planner enables seamless integration of multiple motion modalities within a single path, significantly improving flexibility and adaptability in complex environments. Results demonstrate significantly improved planning performance for 4WIS robots in complex environments.
\end{abstract}


% Please remove or comment out the following line if the footnote is not necessary
% \footnotetext{* This paper will be submitted to Chinese Control Conference 2026.}
\footnotetext{* Corresponding authors: Shoukun Wang (bitwsk@bit.edu.cn).}
\footnotetext{* This work was supported by the National Natural Science Foundation of China under Grant 62473044.}

\section{Introduction}

In recent years, with the continuous advancement of automation and intelligent technologies, mobile robots have gradually expanded beyond the fixed settings of traditional manufacturing, entering increasingly complex and dynamic unstructured environments such as industrial inspection and disaster response \cite{1,2}. In these tasks, conventional steering mechanisms are often limited by large turning radii and restricted motion degrees of freedom, making it difficult to meet the demands for high maneuverability. This has driven the evolution of robotic chassis designs, among which four-wheel independent steering (4WIS) systems have attracted considerable attention. Such systems can operate under multiple steering modes, including Ackermann steering, lateral steering and parallel movement, to adapt to diverse operational requirements \cite{3}. Each mode imposes different constraints on the wheels, resulting in distinct kinematic characteristics \cite{4} and significantly enhancing adaptability and agility in constrained environments.

Owing to these multi-modal motion capabilities, motion planning for 4WIS systems becomes particularly challenging and critical. Motion planning is generally divided into three levels: global path planning, local path planning and path tracking. Existing studies, however, are largely concentrated on the latter two. For example, Wang et al. combined fuzzy logic with deep reinforcement learning to achieve multi-modal local path planning and dynamic decision-making \cite{5}. In addition, various tracking control approaches have been proposed, including backstepping control for symmetric counter-steering \cite{6}, sliding mode control for reverse-steering trajectories \cite{7}, reinforcement learning integrated with model predictive control (MPC) for counter-steering maneuvers \cite{8}, and MPC-based frameworks for stable high-speed trajectory tracking in reverse steering mode \cite{9}, among others. These efforts have substantially advanced the local planning and tracking capabilities of 4WIS robots.

Despite these advances, research on long-range global planning that fully exploits the multi-modal capabilities of 4WIS systems remains limited. Classical planners such as RRT* \cite{10} and A* \cite{11} typically neglect kinematic constraints; optimization-based methods suffer from slow convergence and high computational burden in large-scale scenarios; and learning-based approaches are constrained by memory horizons, which limit their ability to generalize beyond local contexts.

Hybrid A* \cite{13}, which combines graph search with sampling-based strategies, has emerged as an effective approach to balance computational efficiency and scalability while respecting robot kinematics. However, standard Hybrid A* does not explicitly account for the multi-modal motion mechanisms inherent to 4WIS systems, leaving much of their maneuvering potential underutilized. In our previous work \cite{14}, we conducted a preliminary extension of Hybrid A* by incorporating selected 4WIS motion modes into the motion primitive set, thereby enabling rudimentary multi-modal planning. Nevertheless, this approach remained exploratory: the heuristic function was simplified, and issues such as efficient mode switching and integration of deeper kinematic constraints were not systematically addressed.

To overcome these limitations, this paper proposes a systematic multi-modal motion planning framework that extends Hybrid A* by redesigning the heuristic function, optimizing motion primitive generation, and introducing an explicit mode-switching strategy. In particular, the enhanced heuristic accelerates global search by better capturing the kinematic characteristics of four-wheel independent steering robots, while the optimized motion primitives improve path smoothness and adaptability across different maneuvering modes. Furthermore, the proposed mode-switching mechanism explicitly coordinates transitions between steering modes, ensuring both feasibility and efficiency.

\section{Problem Definition}

This section establishes the mathematical framework for global path planning, introduces the kinematic model of four-wheel independent steering robots, and elucidates the implementation principles of four fundamental motion modes.

\subsection{Global Path Planning Problem Formulation}

In the global path planning task, we consider a bounded planar workspace $W \subset \mathbb{R}^2$ containing a finite set of static obstacles $\{O_1, O_2, \ldots, O_m\}$. The obstacle-free region is defined as $W_{\text{free}} = W \setminus \bigcup_{i=1}^m O_i$, representing all regions where the robot can navigate safely.

The mobile robot's state is described by its configuration $\mathbf{x} = (p_x, p_y, \theta)^T$, where $(p_x, p_y) \in \mathbb{R}^2$ denotes the robot's position and $\theta \in [0, 2\pi)$ represents its heading angle. The configuration space is defined as $\mathcal{X} = \mathbb{R}^2 \times \mathbb{S}^1$. At any configuration $\mathbf{x}$, the spatial region occupied by the robot is denoted as $R(\mathbf{x}) \subset W$, which is determined jointly by the robot's geometric shape and current pose.

The collision-free configuration space is mathematically expressed as:
\begin{equation}
\mathcal{X}_{\text{free}} = \{ \mathbf{x} \in \mathcal{X} \mid R(\mathbf{x}) \subseteq W_{\text{free}} \},
\end{equation}
which encompasses all robot configurations that maintain safe clearance from obstacles.

Given an initial configuration $\mathbf{x}_{\text{init}} \in \mathcal{X}_{\text{free}}$ and a goal configuration $\mathbf{x}_{\text{goal}} \in \mathcal{X}_{\text{free}}$, the objective of path planning is to compute a feasible motion trajectory connecting these two states. This trajectory is represented as a parameterized curve $\gamma: [0, 1] \rightarrow \mathcal{X}_{\text{free}}$, satisfying the boundary conditions $\gamma(0) = \mathbf{x}_{\text{init}}$ and $\gamma(1) = \mathbf{x}_{\text{goal}}$. In practical implementations, the continuous trajectory is approximated by a discrete sequence of waypoints:
\begin{equation}
\mathcal{T} = \{\mathbf{x}_i\}_{i=0}^{L},
\end{equation}
where $L$ denotes the number of path segments, with $\mathbf{x}_0 = \mathbf{x}_{\text{init}}$ and $\mathbf{x}_L = \mathbf{x}_{\text{goal}}$.

A valid path must simultaneously satisfy both kinematic feasibility and collision avoidance constraints. Kinematic feasibility requires that transitions between adjacent waypoints comply with the robot's kinematic and dynamic limitations, while collision avoidance ensures that all waypoints belong to the safe configuration space, i.e., $\mathbf{x}_i \in \mathcal{X}_{\text{free}}, \forall i \in \{0, 1, \ldots, L\}$.

\subsection{4WIS Kinematic Model}

\begin{figure}[t]
\centering
\includegraphics[width=0.75\columnwidth]{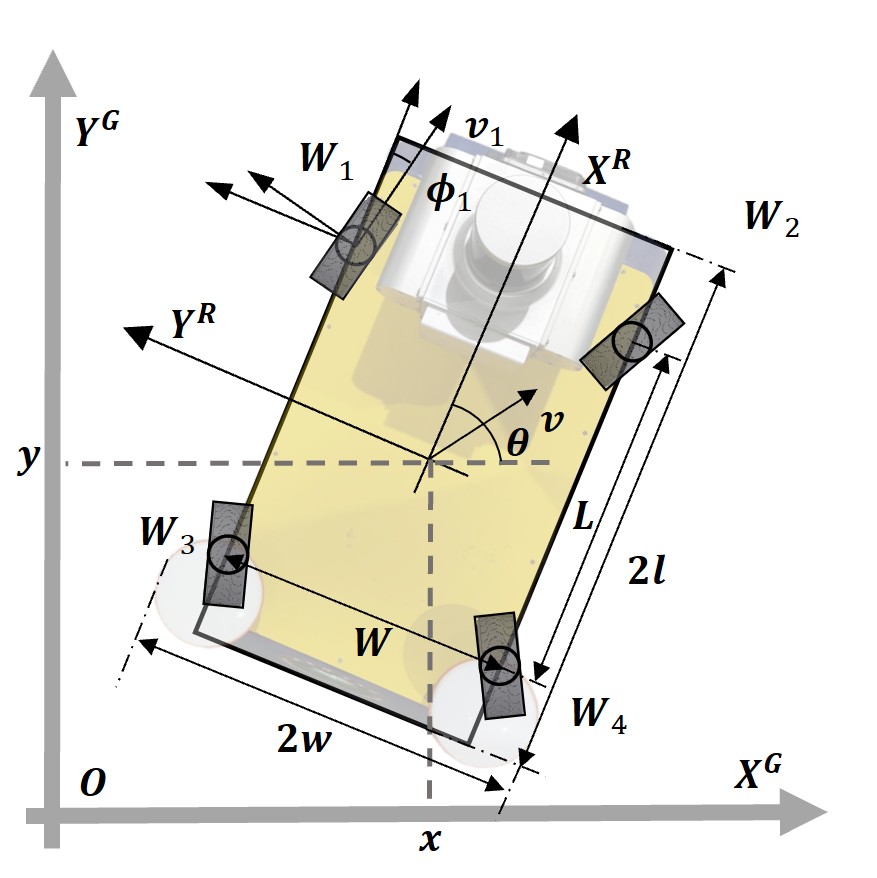}
\caption{Configuration of the 4WIS robot.}
\label{fig1}
\end{figure}

To establish an accurate kinematic description, we make the following reasonable assumptions regarding the robot configuration shown in \cref{fig1}: the center of mass coincides with the geometric center, all four wheels possess identical radii and masses, and slippage effects between tires and ground are neglected. Based on these assumptions, the robot's kinematic equations can be derived.

The position coordinates of the wheels relative to the robot's center are:
\begin{equation}
x^r_i = (-1)^{\lfloor\frac{i-1}{2}\rfloor} L/2, \quad y^r_i = (-1)^{i-1} W/2,
\end{equation}
where $L$ and $W$ represent the wheelbase and track width of the robot, respectively. The kinematic equations governing the robot motion are:
\begin{equation}
\begin{cases}
\dot{x} = \frac{1}{4}\sum_{i=1}^{4}v\cos(\phi_i+\theta)\\
\dot{y} = \frac{1}{4}\sum_{i=1}^{4}v\sin(\phi_i+\theta)\\
\dot \theta =\sum_{i=1}^{4} \frac{v(-y^r_i\cos\phi_i+x^r_i\sin\phi_i)}{4{x^r_i}^2+4{y^r_i}^2},
\end{cases}
\end{equation}
where $v_i$ and $\phi_i$ denote the linear velocity and steering angle of the $i$-th wheel. The subscripts 1, 2, 3, and 4 correspond to the front-left, front-right, rear-left, and rear-right wheels, respectively.

\subsection{4WIS Motion Modes}

The four-wheel independent steering system supports three fundamental motion modes: Ackermann steering, lateral steering and parallel movement. These modes are employed for path following, lateral maneuvering, omnidirectional movement, and zero-radius turning, respectively, providing the foundation for flexible navigation in complex environments.

Ackermann steering achieves smooth cornering by directing all wheels toward a common instantaneous center of rotation. For given control inputs $\{v, \phi\}$, the relationships between wheel steering angles and velocities are:
\begin{equation}
\begin{cases}
\cot\phi_1 = -\cot\phi_3 = \cot\phi - W/L\\
\cot\phi_2 = -\cot\phi_4 = \cot\phi + W/L\\
v_1 = v_3 = v\tan\phi/\sin\phi_1\\
v_2 = v_4 = v\tan\phi/\sin\phi_2,
\end{cases}
\label{motion1}
\end{equation}
where $v$ denotes the robot's longitudinal velocity, $\phi$ represents the desired steering angle, $\phi_i$ is the steering angle of the $i$-th wheel, and $v_i$ is the linear velocity of the $i$-th wheel. 

Lateral steering preserves the same same geometric constraint principles as the Ackermann mode, but the wheels are reoriented to be perpendicular to the robot’s longitudinal axis, allowing the robot to generate pure sideways motion. When the lateral direction is regarded as the effective forward direction, the underlying steering-angle computation becomes structurally identical to that of the Ackermann formulation, with $L$ and $W$ exchanging roles in the geometric constraint. In parallel movement mode, all wheels maintain identical steering angles and velocities, so that the robot undergoes pure translation in an arbitrary direction without changing its orientation.

% When the four wheels maintain the same velocity, for the control input angle $\phi$, the motion constraints are:
% \begin{equation}
% \begin{cases}
% \phi_1 = \phi_3 = \phi\\
% \phi_2 = \phi_4 = \text{sgn}(\phi)(\pi - |\phi|),
% \end{cases}
% \label{motion2}
% \end{equation}
% where $\text{sgn}(\cdot)$ denotes the sign function.

% The in-place rotation mode enables zero-radius turning, with each wheel's steering angle uniquely determined by the robot's geometric parameters:
% \begin{equation}
% \begin{cases}
% \phi_1 = -\phi_3 = \pi - \arctan\left(L/W\right)\\
% \phi_2 = -\phi_4 = \arctan\left(L/W\right),
% \end{cases}
% \label{motion3}
% \end{equation}
% with all four wheels also need to maintain the same speed.

\section{Methodology}

\subsection{Review of the Hybrid A* Algorithm}

The Hybrid A* algorithm is a path planning method specifically designed for mobile robots subject to kinematic constraints. This algorithm introduces continuous state space motion feasibility reasoning within a discrete grid search framework, thereby generating trajectories that satisfy both kinematic constraints and practical feasibility. Nodes are described by position $(x, y)$, orientation $\theta$, motion direction $d$, steering angle $\delta$, and accumulated cost $g(n)$, where $d \in \{+1, -1\}$ represents forward and reverse motion, respectively. The accumulated cost consists of path length and multiple penalty terms:
\begin{align}
g(n) & = g_{\text{prev}} + \Delta s 
       + C_{\text{reverse}} \cdot \mathbb{I}_{\{d = -1\}} 
       + C_{\text{steer}} \cdot |\delta| \notag\\
     &+ C_{\text{steer-change}} \cdot |\Delta \delta| 
       + C_{\text{direction-change}} \cdot \mathbb{I}_{\{d \neq d_{\text{prev}}\}},
\end{align}
where $g_{\text{prev}}$ denotes the cost of the parent node, $\Delta s$ denotes the incremental path length, $\mathbb{I}$ denotes the indicator function for reverse motion and direction changes, and $\Delta \delta$ denotes the change in steering angle. Each $C$ term corresponds to the configurable weight of its respective penalty term.

The total evaluation function for a node is defined as:
\begin{equation}
f(n) = g(n) + h(n),
\end{equation}
where $h(n)$ is the heuristic cost. To balance efficiency and quality, Hybrid A* employs a dual heuristic strategy: on one hand, the Euclidean distance transform pre-computed based on Dijkstra or A* can reflect the shortest path estimate when considering obstacle constraints but ignoring kinematic constraints; on the other hand, Reeds-Shepp (RS) curves provide optimal cost estimates when ignoring obstacle constraints but strictly satisfying robot kinematic constraints. Taking the maximum of both as the heuristic maintains admissibility while effectively guiding the search.

When approaching the target, the algorithm attempts to connect directly to the goal through RS curves. If the trajectory is collision-free, the complete path is returned; otherwise, conventional expansion continues. Finally, the algorithm obtains a complete path that is both feasible and collision-free by backtracking through parent node pointers.

% All generated trajectories require collision detection, typically based on KDTree or grid maps to calculate the minimum distance between trajectory points and obstacles. If below a safety threshold, a collision is determined. 
\subsection{Multi-Modal State Representation}

Traditional Hybrid A* algorithms are designed for robots with fixed motion patterns, typically based on Ackermann steering models, and cannot fully exploit the multi-modal motion advantages of four-wheel independent steering systems. To overcome the limitations of traditional algorithms, this research enhances state representation and search strategies to achieve unified modeling of multiple motion modes.

% Four-wheel independent steering and drive systems can achieve multiple maneuvering modes including Ackermann steering, lateral steering, parallel movement, and in-place rotation, providing stronger adaptability and flexibility in complex environments. 

The core idea is to extend the original three-dimensional state space $(x, y, \theta)$ to a four-dimensional state space $(x, y, \theta, m)$, where the motion mode variable $m$ is explicitly modeled as the fourth dimension of the state. Specifically, $m \in \{1, 2, 3\}$ corresponds to Ackermann steering, lateral steering and parallel movement, respectively. 

% Constraint formulas are given in \cref{motion1,motion2,motion3}.

\begin{figure}[t]
\centering
\includegraphics[width=\columnwidth]{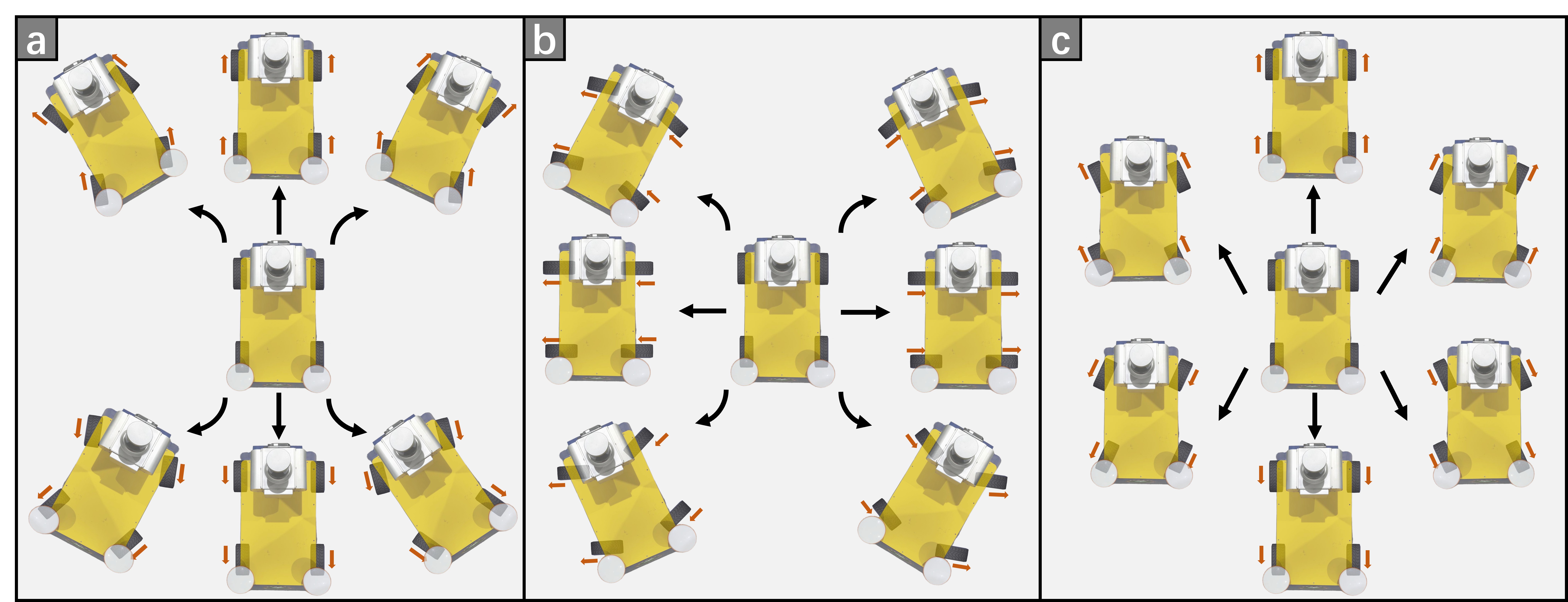}
\caption{Multi-modal motion sampling. (a) Ackermann steering. (b) Lateral steering. (c) Parallel movement.}
\label{fig2}
\end{figure}

In the expanded four-dimensional state space, state expansion process includes two operations: intra-modal expansion and inter-modal expansion. Intra-modal expansion maintains the current motion mode unchanged and generates successor states in the three-dimensional pose space according to the corresponding kinematic model; inter-modal expansion performs mode switching at the current position, producing successor states with different motion modes but the same pose. To effectively manage state space complexity, the algorithm employs heuristic modal selection strategies, choosing the most effective motion mode based on the relative state between the current position and target, and minimizing unnecessary modal switches while maintaining trajectory continuity under constraint satisfaction. The sampling strategies associated with each motion mode are illustrated in \cref{fig2}.

To accommodate the needs of multi-modal expansion, we introduce a reference velocity and sampling time mechanism to replace the traditional fixed forward iteration distance method. By setting reference velocity $v_{\text{ref}}$ and iteration time $\Delta t$, the forward iteration distance is dynamically calculated as $\Delta s = v_{\text{ref}} \times \Delta t$. This improvement lays the foundation for subsequent multi-modal RS curve expansion and mode switching cost quantification.

\subsection{Multi-Modal Reeds-Shepp Curves}

To accommodate the multi-modal motion characteristics of four-wheel independent steering systems, this research extends the traditional RS curve calculation method. The generation of RS curves depends on key parameters such as maximum curvature, and different motion modes have different kinematic characteristics, thus requiring separate design of corresponding curve calculation methods.

% In Ackermann steering mode, the RS curve is consistent with traditional algorithms, with minimum turning radius:
% \begin{equation}
% R_{\min}^{(1)} = \frac{L}{\tan(\delta_{\max})},
% \end{equation}
% where $\delta_{\max}$ is the maximum front wheel steering angle.

In Ackermann steering mode, the RS curve is consistent with traditional algorithms, with maximum curvature:
\begin{equation}
\kappa^{(1)} = 2\tan(\phi_{\max})/{L},
\end{equation}
where $\phi_{\max}$ is the maximum wheel steering angle.

In lateral steering mode, the algorithm converts the orientation from the robot's forward direction to the right direction and uses the track width $W$ as the equivalent wheelbase:
\begin{equation}
\kappa^{(2)} = 2\tan(\phi_{\max})/{W}.
\end{equation}
The corresponding RS curves require coordinate transformation to adapt to lateral motion characteristics.

\begin{figure}[t]
\centering
\includegraphics[width=\columnwidth]{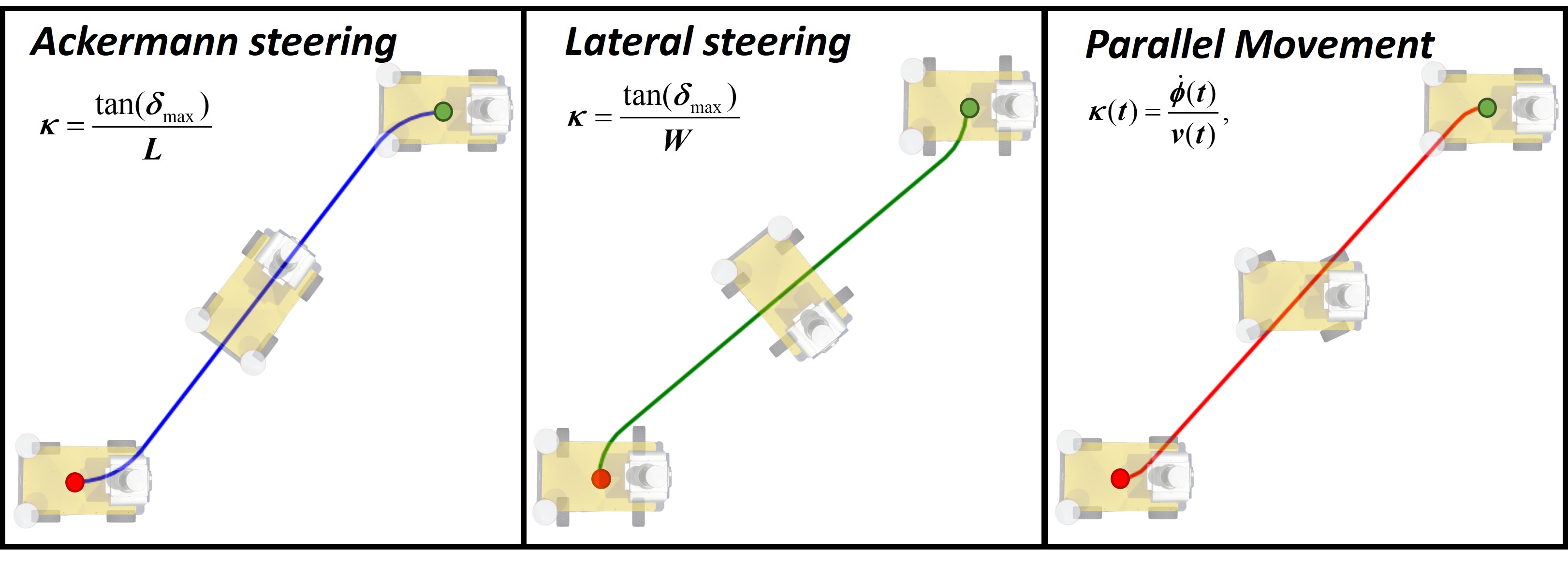}
\caption{Visualization of multi-Modal Reeds-Shepp curves.}
\label{fig3}
\end{figure}

In parallel movement mode, the robot's orientation angle remains unchanged, but the center of mass trajectory can still form curves. When the wheel angle $\delta$ changes with time, the trajectory curvature depends on the ratio of the steering angle change rate to the translation velocity, expressed as:
\begin{equation}
\kappa^{(3)}(t) = \dot{\phi}(t)/{v(t)},
\end{equation}
where $\dot{\phi}(t)$ is the rate of change of wheel angle and $v(t)$ is the reference velocity. This expression indicates that, unlike Ackermann mode where curvature is determined solely by geometric parameters, curvature in parallel mode is directly coupled with velocity: the same wheel angle change rate produces larger curvature at low speeds and corresponds to gentler trajectories at high speeds. This makes parallel movement mode a special case in RS curve extension. Since the robot's orientation remains constant in this mode, it is only activated when the current orientation aligns with the target orientation or falls within a specified tolerance. The Reeds–Shepp curve visualizations for all motion modes are shown in \cref{fig3}.

\subsection{Cost and Heuristic Functions}

Based on the multi-modal RS curve design, we construct a heuristic function adapted to the four-dimensional state space. Traditional hybrid A* uses the maximum of Euclidean distance and RS curve cost as the heuristic function, but suffers from underestimation problems, particularly when obstacle circumvention or complex maneuvers are required. In multi-modal scenarios, this problem becomes more prominent as the algorithm needs to predict possible mode switching sequences and associated costs.

% The cost function is extended by introducing a mode switching penalty into the traditional $g(n)$. The calculation of mode switching cost directly utilizes the aforementioned reference velocity and sampling time mechanism: $C_{\text{switch}} = v_{\text{ref}} \times t_{\text{switch}}$, where $t_{\text{switch}}$ is the preset switching time. This time-velocity based quantification method enables mode switching costs to be compared with path length in the same dimension. The extended accumulated cost function is expressed as:

The cost function is extended by introducing a mode switching penalty into the traditional $g(n)$. The calculation of the mode switching cost leverages the aforementioned reference velocity and sampling time mechanism and explicitly accounts for the required full stop and subsequent re-acceleration during a mode transition:
$$C_{\text{switch}} = v_{\text{ref}} \times t_{\text{switch}} + v_\text{ref}^2/a_{max},$$ 
where $t_{\text{switch}}$ denotes the preset switching time and $a$ represents the robot acceleration. For mode switching occurring at the initial node, the system starts from rest and only undergoes the acceleration phase. As a result, the acceleration-related component of the switching cost is halved, since no deceleration process is required. This time-velocity based quantification method enables mode switching costs to be compared with path length in the same dimension. The extended accumulated cost function is expressed as:
\begin{align}
& g(n)  = g_{\text{prev}} + \Delta s 
       + C_{\text{reverse}} \cdot \mathbb{I}_{\{d = -1\}} 
       + C_{\text{steer}} \cdot |\phi| \notag\\
     & + C_{\text{steer-change}} \cdot |\Delta \phi| \notag\\
    & +  C_{\text{direction-change}} \cdot \mathbb{I}_{\{d \neq d_{\text{prev}}\}} + C_{\text{switch}} \cdot \mathbb{I}_{\{m \neq m_{\text{prev}}\}}.
\end{align}

It should be noted that at mode switching nodes, all cost terms except the mode switching cost are considered zero, as these cost terms are designed only to evaluate state iterations within the same motion mode.

To address the underestimation problem of traditional heuristic functions and adapt to multi-modal characteristics, this research designs an enhanced heuristic function based on multi-modal RS curves:
\begin{align}
\!\! h(n) = \max\{h_{\text{euc}}, \min_{m' \in M} [h_{\text{RS}}^{(m')}+ C_{\text{switch}} \cdot \mathbb{I}_{\{m \neq m'\}}]\},\!\!
\end{align}
where $h_{\text{euc}}$ is the unconstrained heuristic term based on Euclidean distance, $h_{\text{RS}}^{(m')}$ is the RS curve heuristic term based on corresponding kinematic constraints in mode $m'$, and $M = \{1, 2, 3\}$ is the set of all available motion modes. This design provides more accurate cost estimates while maintaining heuristic function admissibility.

\begin{algorithm}[t]
\caption{Extended Hybrid A* for Multi-Modal Planning}
\label{alg:multimodal_hybrid_astar}
\begin{algorithmic}[1]
\State Initialize open list with start node $(x_0, y_0, \theta_0, m_0)$
\State Initialize closed list as empty
\While{open list is not empty}
    \State $n \leftarrow$ node with minimum $f(n)$ from open list
    \If{$n$ is close to goal}
        \State Compute all multi-modal RS curves to goal
        \State Sort RS curves by ascending connection cost 
        \For{each RS curve in sorted order}
            \If{RS curve is collision-free}
                \State \Return path 
            \EndIf
        \EndFor
    \EndIf
    \State Move $n$ to closed list
    \For{each available motion mode $m'$}
        \If{$m' = m_{\text{current}}$}
            \State Generate intra-modal successors
        \Else
            \State Generate inter-modal successors (mode switch)
        \EndIf
        \State Compute $g(n')$ with mode switching cost
        \State Compute multi-modal heuristic $h(n')$
        \If{$n'$ is collision-free and not in closed list}
            \State Add $n'$ to open list
        \EndIf
    \EndFor
\EndWhile
\State \Return failure
\end{algorithmic}
\end{algorithm}

\subsection{Terminal Connection Strategy}

The algorithm begins attempting direct connection using multi-modal RS curves when approaching the target within a certain distance. Connection strategies include both RS curve connection using the current mode and RS curve connection after switching to other modes. For connections involving mode switching, the algorithm must additionally consider mode switching costs. Among all feasible connection schemes, the algorithm selects the curve with minimum total cost as the terminal path for rapid connection. The total connection cost is calculated as:
\begin{equation}
C_{\text{connection}} = C_{\text{RS-curve}} + C_{\text{switch}} \cdot \mathbb{I}_{\{m_{\text{current}} \neq m_{\text{connection}}\}},
\end{equation}
where $C_{\text{RS-curve}}$ is the cost of the RS curve itself, $m_{\text{current}}$ is the motion mode of the current node, and $m_{\text{connection}}$ is the motion mode used for connection. This terminal connection strategy ensures both kinematic feasibility of the path and optimizes connection efficiency through mode selection. \cref{alg:multimodal_hybrid_astar} shows the entire procedure of extended hybrid A*.

% It is worth noting that paths obtained through hybrid A* as a front-end planner typically still require back-end optimization processing to further improve smoothness, safety, and dynamic feasibility. Back-end optimization can be achieved through various methods, such as numerical optimization-based trajectory smoothing techniques, interpolation algorithms, or predictive correction strategies based on robot dynamics models. However, given that this paper focuses on multi-modal path generation problems for four-wheel independent steering robots, our discussion primarily concentrates on extensions and improvements to the front-end hybrid A* algorithm. Due to the introduction of multiple motion modes, back-end optimization not only needs to handle optimization of path geometric properties but also requires additional consideration of reasonableness detection and sequence optimization of mode switching points. This problem extends beyond the current research scope and will be explored in depth as an important direction for future work.

It should be noted that paths generated by hybrid A* typically require back-end optimization for smoothness, safety, and dynamic feasibility. While such optimization can be performed using trajectory smoothing, interpolation, or predictive correction, this paper focuses on multi-modal path generation for four-wheel independent steering robots, emphasizing front-end hybrid A* extensions. Handling back-end optimization with multiple motion modes, including mode-switch validation and sequence adjustment, is beyond the current scope and will be addressed in future work.

\section{Experiments}

\subsection{Experimental Setup}

To validate the proposed multi-modal Hybrid A* framework, we conducted simulations on a computing platform equipped with an Intel i7-13700H CPU and an NVIDIA RTX 4060 GPU. The simulated robot adopts a four-wheel independent steering configuration, and its key kinematic and geometric parameters are summarized in \cref{tab1}. 

\begin{table}[h]
\centering
\caption{Key Parameters of the 4WIS Robot}
\label{tab1}
\scalebox{0.82}{ % 缩放到85%
\begin{tabular}{l c l c}
\toprule
\textbf{Parameter (Symbol)} & \textbf{Value} & \textbf{Parameter (Symbol)} & \textbf{Value} \\
\midrule
Length ($2l$) & 1.00 m & Width ($2w$) & 0.62 m \\
Wheelbase ($L$) & 0.68 m & Track Width ($W$) & 0.52 m \\
Max Steer Angle ($\phi_{\max}$) & 30° & Max Steer Rate ($\dot\phi_{\max}$) & 360°/s \\
Reference Velocity ($v_{\text{ref}}$) & 1.0 m/s & Sampling Time ($\Delta t$) & 0.2 s \\
Mode Switch Time ($t_{\text{switch}}$) & 0.5 s & Max Acceleration ($a_{\max}$) & 2 $\text{m/s}^2$\\
\bottomrule
\end{tabular}}
\end{table}

Two representative environments were designed to evaluate the planner under different conditions. The first environment is a 20 m × 20 m maze with a corridor width of 2 m, primarily used to simulate planning challenges in narrow passages and frequent turns. The second environment is a structured parking scenario, including lanes and parking slots, which is intended to test maneuverability and precision in a structured space. For each environment, three test cases of varying complexity were constructed. The main planning parameters of the Hybrid A* algorithm are listed in \cref{tab2}.

\begin{table}[h]
\centering
\caption{Key Planning Parameters of Hybrid A*}
\label{tab2}
\scalebox{0.85}{ % 缩放到85%
\begin{tabular*}{\columnwidth}{l@{\extracolsep{\fill}}lc}
\toprule
\textbf{Parameter} & \textbf{Symbol} & \textbf{Value} \\
\midrule
Reverse Motion Penalty & $C_{\text{reverse}}$ & 2 \\
Steering Angle Penalty & $C_{\text{steer}}$ & 1 \\
Steering Change Penalty & $C_{\text{steer-change}}$ & 1 \\
Direction Change Penalty & $C_{\text{direction-change}}$ & 1 \\
\bottomrule
\end{tabular*}}
\end{table}

For performance evaluation, path length and path cost were selected as the primary metrics. Path length reflects geometric efficiency, while path cost incorporates path length, the number of reverse motions, steering angle variations, and mode-switching penalties, thereby providing a more comprehensive assessment of planning quality.

\subsection{Performance Comparison}

\begin{figure}[t]
\centering
\includegraphics[width=\columnwidth]{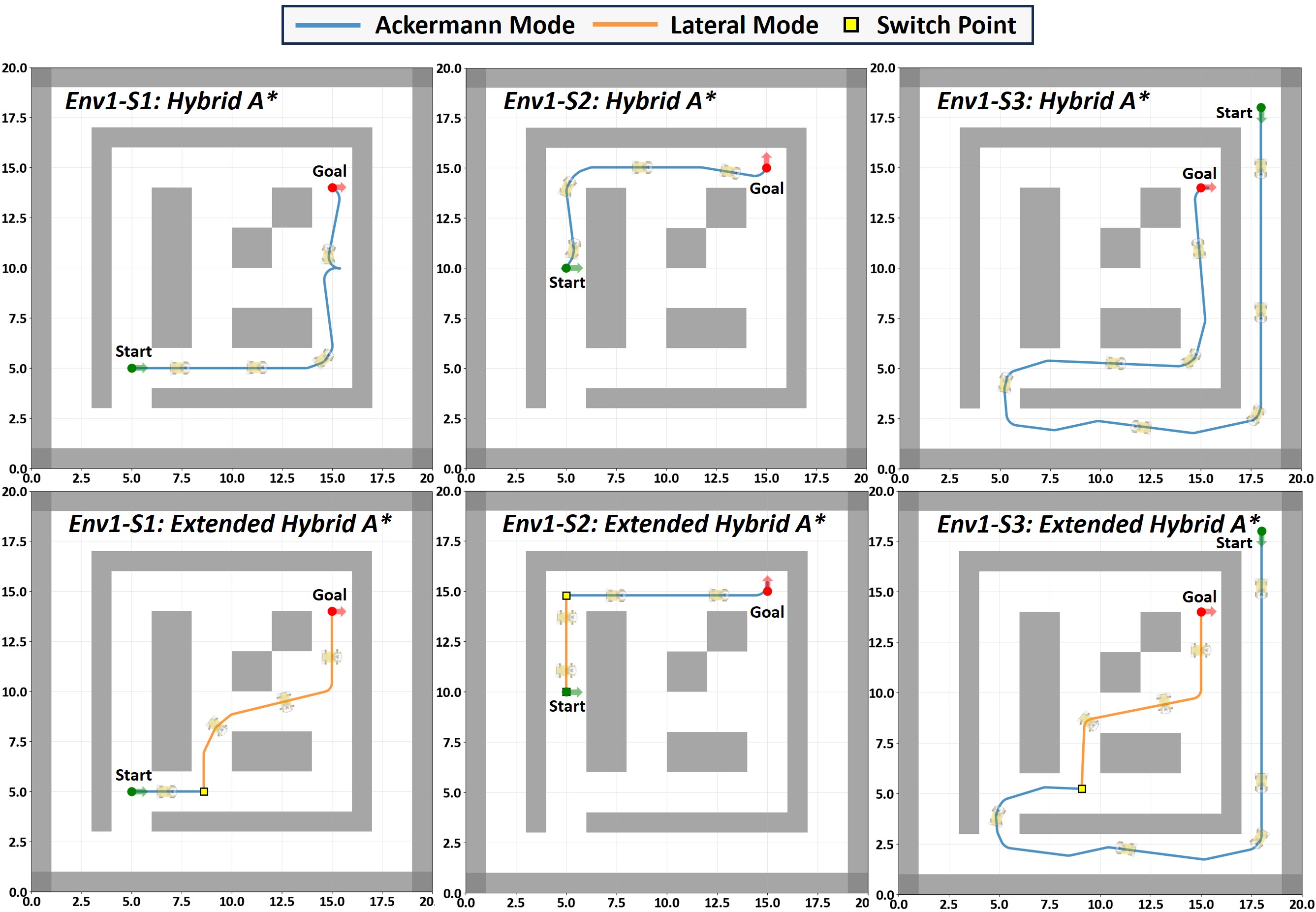}
\caption{Path planning results in maze environment using baseline Hybrid A* and the proposed extended Hybrid A*.}
\label{fig4}
\end{figure}

\begin{figure}[t]
\centering
\includegraphics[width=\columnwidth]{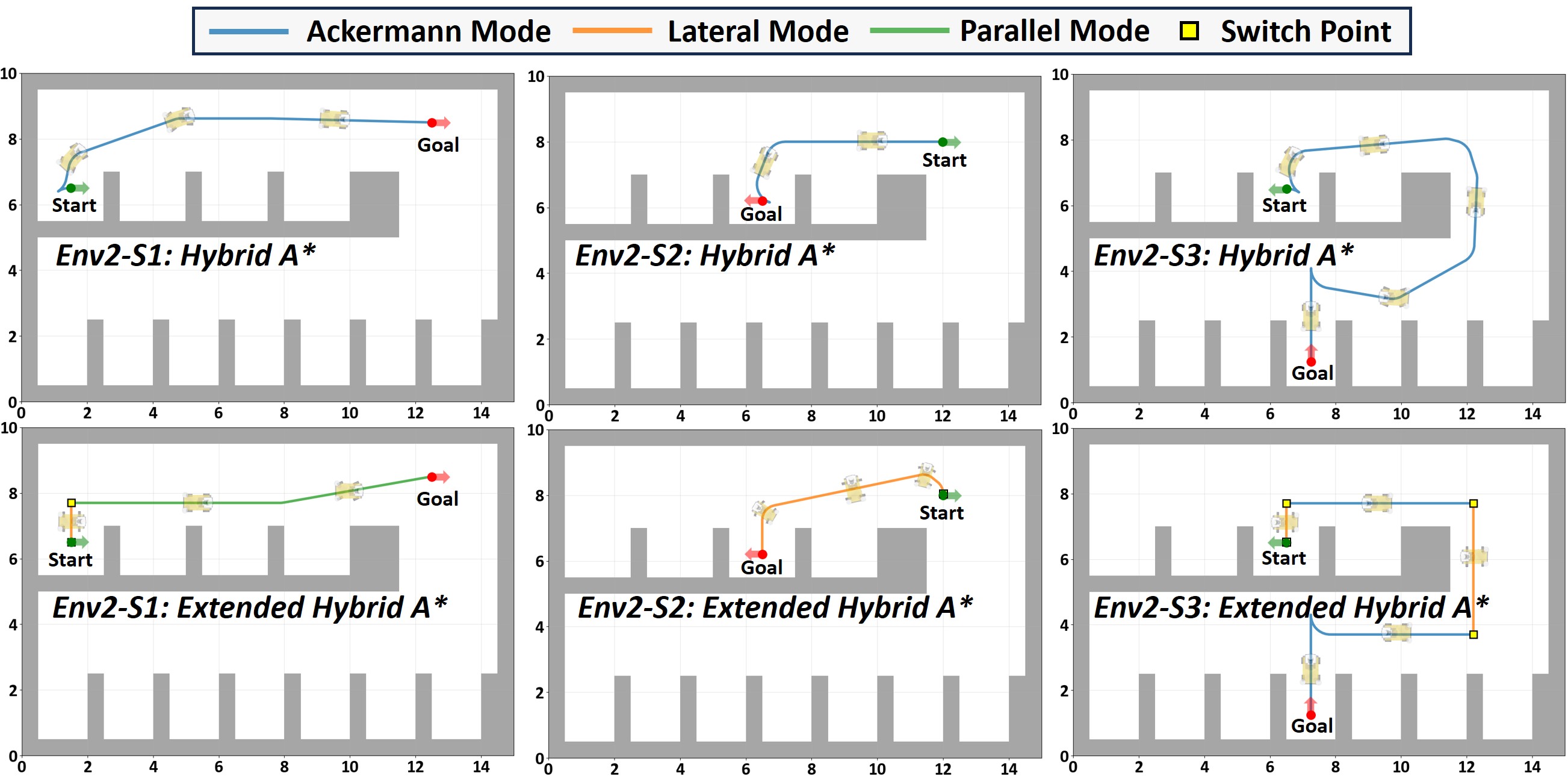}
\caption{Path planning results in parking environment using baseline Hybrid A* and the proposed extended Hybrid A*.}
\label{fig5}
\end{figure}

In the maze environment, the comparative results are shown in \cref{fig4}. Due to the presence of narrow corridors and sharp turns, the baseline Hybrid A* relying solely on Ackermann steering often produces longer and more tortuous paths. In contrast, the proposed multi-modal Hybrid A* can flexibly switch motion modes according to the scenario. For example, in Env1-S1 and Env1-S2, the lateral mode enables the robot to directly traverse narrow passages, resulting in shorter and smoother paths. In the more challenging Env1-S3, the planner seamlessly integrates Ackermann and lateral maneuvers, effectively reducing steering costs while maintaining path feasibility.

In the parking environment, the comparison is illustrated in \cref{fig5}. For side-parking and exit tasks (Env2-S1 and Env2-S2), the baseline Hybrid A* frequently requires multiple adjustments and backward maneuvers, whereas the multi-modal planner leverages lateral and translational movements to achieve more efficient parking operations. In the long-distance slot-changing scenario (Env2-S3), the extended planner further demonstrates the advantage of motion mode combinations, significantly reducing unnecessary reversing and large steering maneuvers. This capability highlights its superior mobility and operational efficiency in structured environments.

\cref{tab:performance_comparison} summarizes the quantitative differences between the two approaches. Across all test cases, the proposed method consistently outperforms the baseline in terms of both path length and path cost, with the most significant improvements observed in cost reduction. This is attributed to the cost function explicitly penalizing excessive steering and reversing, while the multi-modal switching strategy effectively mitigates these penalties. Overall, the results confirm that the proposed framework not only enhances geometric efficiency but also demonstrates strong generality and robustness across diverse tasks, providing practical potential for 4WIS robots operating in complex environments.

% \begin{table}[t]
% \centering
% \caption{Performance Comparison between Hybrid A* and Extended Multi-modal Hybrid A* in Various Scene}
% \label{tab:performance_comparison}
% \scalebox{0.85}{ % 缩放到85%
% \begin{tabular}{c l c c}
% \toprule
% \textbf{Scenario} & \textbf{Algorithm} & \textbf{Path Length (m)} & \textbf{Path Cost} \\
% \midrule
% \multirow{2}{*}{Env1-S1} & Hybrid A* & 18.49 & 34.75 \\
%                     & Multi-moda Hybrid A* & \textbf{15.88} & \textbf{20.03} \\
% \midrule
% \multirow{2}{*}{Env1-S2} & Hybrid A* & 14.89 & 25.89 \\
%                     & Multi-modal Hybrid A* & \textbf{14.80} & \textbf{19.44} \\
% \midrule
% \multirow{2}{*}{Env1-S3} & Hybrid A* & 48.56 & 59.78 \\
%                     & Multi-modal Hybrid A* & \textbf{45.35} & \textbf{58.08} \\
% \midrule
% \multirow{2}{*}{Env2-S1} & Hybrid A* & 12.69 & 18.48 \\
%                     & Multi-modal Hybrid A* & \textbf{12.27} & \textbf{15.32} \\
% \midrule
% \multirow{2}{*}{Env2-S2} & Hybrid A* & 6.73 & 16.04 \\
%                     & Multi-modal Hybrid A* & \textbf{6.44} & \textbf{9.87} \\
% \midrule
% \multirow{2}{*}{Env2-S3} & Hybrid A* & 16.61 & 34.76 \\
%                     & Multi-modal Hybrid A* & \textbf{16.49} & \textbf{33.69} \\
% \bottomrule
% \end{tabular}}
% \end{table}

\begin{table}[t]
\centering
\caption{Performance Comparison between Hybrid A* and Multi-modal Hybrid A*}
\label{tab:performance_comparison}
\scalebox{0.85}{ % 缩放到85%
\small
\begin{tabular}{l cc cc}
\toprule
\multirow{2}{*}{\textbf{Scenario}}
& \multicolumn{2}{c}{\textbf{Hybrid A*}} 
& \multicolumn{2}{c}{\textbf{Multi-modal Hybrid A*}} \\
\cmidrule(lr){2-3} \cmidrule(lr){4-5}
& \textbf{Path Length} & \textbf{Path Cost} & \textbf{Path Length} & \textbf{Path Cost} \\
\midrule
Env1-S1 & 18.49 & 34.75 & \textbf{15.88} & \textbf{20.03} \\
Env1-S2 & 14.89 & 25.89 & \textbf{14.80} & \textbf{19.44} \\
Env1-S3 & 48.56 & 59.78 & \textbf{45.35} & \textbf{58.08} \\
Env2-S1 & 12.69 & 18.48 & \textbf{12.27} & \textbf{15.32} \\
Env2-S2 & 6.73  & 16.04 & \textbf{6.44}  & \textbf{9.87} \\
Env2-S3 & 16.61 & 34.76 & \textbf{16.49} & \textbf{33.69} \\
\bottomrule
\end{tabular}}
\end{table}

\begin{figure}[t]
\centering
\includegraphics[width=0.7\columnwidth]{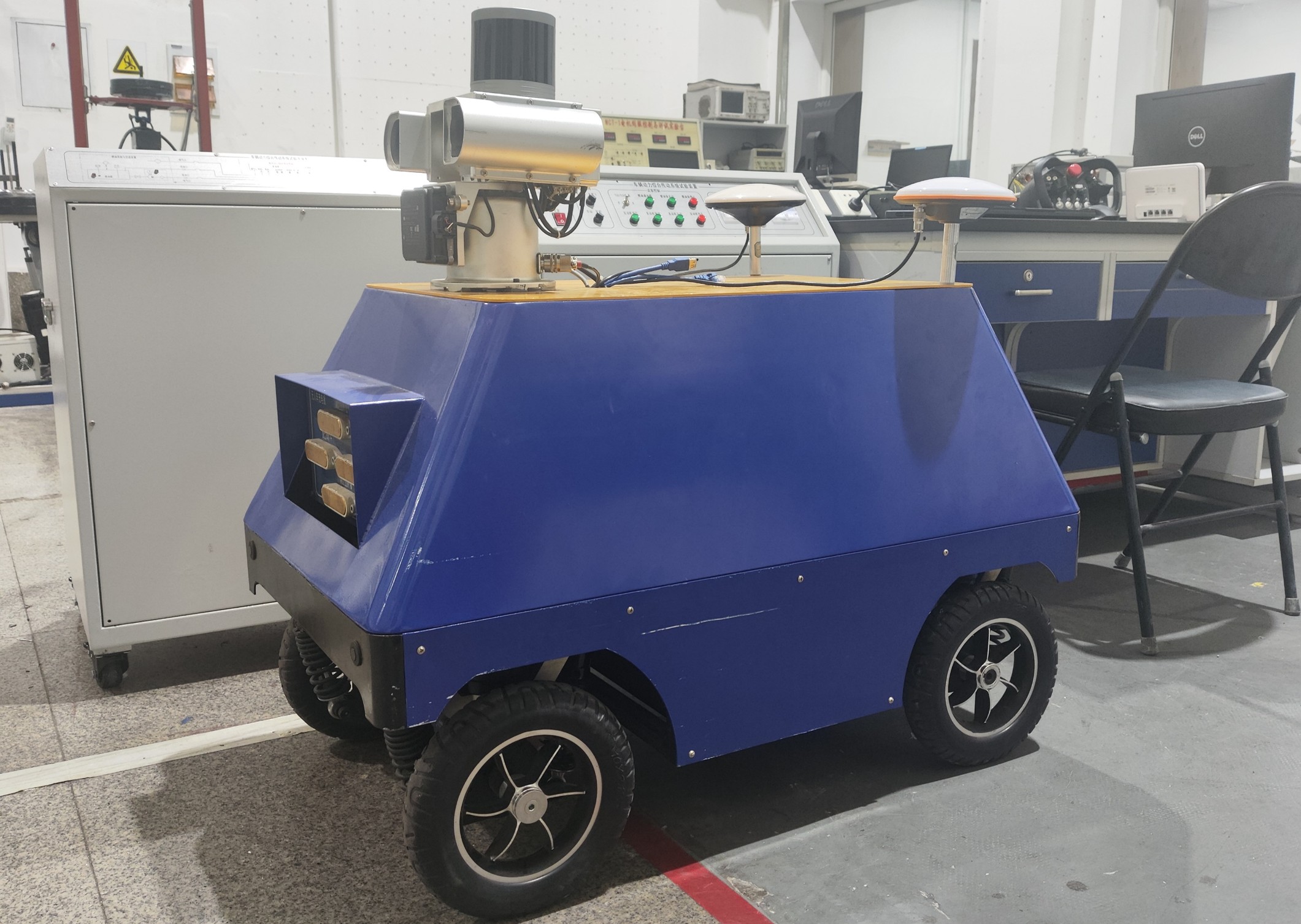}
\caption{Physical 4WIS Platform.}
\label{fig6}
\end{figure}

To further validate the feasibility of the proposed framework in practical applications, we conducted preliminary experiments on a physical 4WIS platform as shown in \cref{fig6}. DRL tracking controllers \cite{15} were constructed for different modes, and tracking tests were performed on six multi-modal trajectories planned in simulation scenarios. The selected test trajectories cover most typical maneuvering actions and mode switching scenarios, demonstrating good representativeness. Under a reference velocity of 1 m/s, the average lateral error was controlled within 3 cm and the average longitudinal error was kept below 5 cm, including errors at switching points, which validates the feasibility and effectiveness of the planned trajectories in real systems.

\section{Conclusion}

This paper presented a multi-modal Hybrid A* framework for four-wheel independent steering robots. By extending the state space to explicitly model motion modes and developing corresponding multi-modal RS curves, the algorithm fully exploits 4WIS motion capabilities. The enhanced heuristic function and terminal connection strategy ensure optimal path generation while maintaining computational efficiency. Experimental validation demonstrates the effectiveness of the proposed approach.

Future work will focus on back-end optimization for multi-modal trajectories, including trajectory smoothing techniques that preserve mode switching points, sequence optimization methods for refining mode transition decisions, and integration of dynamic feasibility constraints. Further validation on physical platforms will also be conducted to evaluate real-world performance.

\end{document}